\documentclass{article}
\usepackage{spconf,amsmath,graphicx}

\usepackage{amsbsy}
\usepackage{bm}
\usepackage[T1]{fontenc}
\usepackage{booktabs} 
\usepackage{multirow}

\usepackage{cite}
\usepackage{amssymb}
\usepackage[caption = false, font=footnotesize]{subfig}
\usepackage{float}

\usepackage{array}
\usepackage{color}
\usepackage{tabularx}
\usepackage{longtable}
\usepackage{tabu}

\usepackage{array}
\usepackage{color}
\usepackage{tabularx}
\usepackage{longtable}
\usepackage{tabu}
\usepackage{multirow}

\newcommand{\ie}{\textit{i}.\textit{e}.}

\title{Learning to Blindly Assess Image Quality in the Laboratory and Wild}
\name{Weixia Zhang\textsuperscript{*}, Kede Ma\textsuperscript{\dag}, Guangtao Zhai\textsuperscript{*}, Xiaokang Yang\textsuperscript{*}\thanks{This work was supported in part by the National Natural Science Foundation of China under Grant 61901262 and CCF-Tecent Rhino-Bird Young Faculty Open Research Fund.}}
\address{\textsuperscript{*} Artificial Intelligence Institute, Shanghai Jiao Tong University\\
\textsuperscript{\dag} Department of Computer Science, City University of Hong Kong}
\begin{document}
\ninept
\maketitle
\begin{abstract}
Computational models for blind image quality assessment (BIQA) are typically trained in well-controlled laboratory environments with limited generalizability to realistically distorted images. Similarly, BIQA models optimized for images captured in the wild cannot adequately handle synthetically distorted images. To face the cross-distortion-scenario challenge, we develop a BIQA model and an approach of training it on multiple IQA databases (of different distortion scenarios) simultaneously. A key step in our approach is to create and combine image pairs within individual databases as the training set, which effectively bypasses the issue of perceptual scale realignment. We compute a \textit{continuous} quality annotation for each pair from the corresponding human opinions, indicating the \textit{probability} of one image having better perceptual quality. We train a deep neural network for BIQA over the training set of massive image pairs by minimizing the fidelity loss. Experiments on six IQA databases demonstrate that the optimized model by the proposed training strategy is effective in blindly assessing image quality in the laboratory and wild, outperforming previous BIQA methods by a large margin.
\end{abstract}
\begin{keywords}
    Blind image quality assessment, deep neural networks, database combination, fidelity loss.
\end{keywords}
\section{Introduction}
\label{sec:intro}

Blind image quality assessment (BIQA) aims to predict the perceived quality of a visual image without reference to its original pristine-quality counterpart. The majority of BIQA models~\cite{mittal2012no,ye2012unsupervised,ma2018end,bosse2018deep} have been developed in well-controlled laboratory environments, whose feature representations have been adapted to common synthetic distortions (\textit{e.g.}, Gaussian blur and JPEG compression). Only recently has BIQA of realistically distorted images captured in the wild become an active research topic~\cite{ghadiyaram2016massive}. Poor lighting conditions, sensor limitations, lens imperfections, and amateur manipulations are the main sources of distortions in this scenario, which are generally more complex and difficult to simulate. As a result, BIQA models trained on databases of synthetic distortions (\textit{e.g.}, LIVE~\cite{sheikh2006statistical} and TID2013~\cite{Ponomarenko201557}) are not capable of handling databases of realistic distortions (\textit{e.g.}, LIVE Challenge~\cite{ghadiyaram2016massive} and KonIQ-10K~\cite{koniq10k}). Similarly, models optimized for realistic distortions do not work well for synthetic distortions~\cite{zhang2019deep}.

\begin{figure}[t]
    \centering
    \captionsetup{justification=centering}
    \subfloat[]{\includegraphics[width=0.23\textwidth]{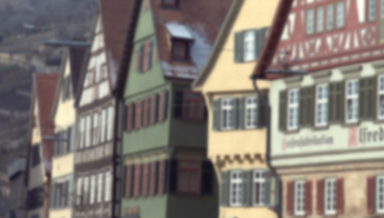}}\hskip.2em
    \subfloat[]{\includegraphics[width=0.23\textwidth]{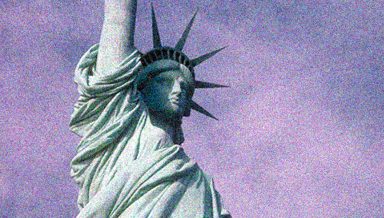}}\hskip.2em
    \subfloat[]{\includegraphics[width=0.23\textwidth]{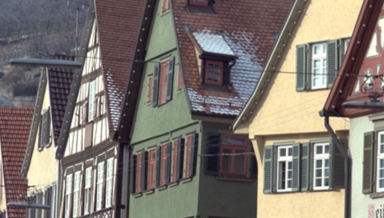}}
    \vspace{0em}
    \subfloat[]{\includegraphics[width=0.23\textwidth]{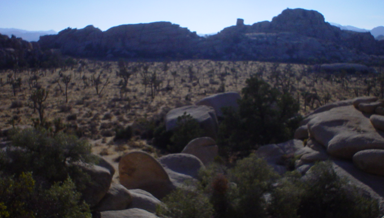}}\hskip.2em
    \subfloat[]{\includegraphics[width=0.23\textwidth]{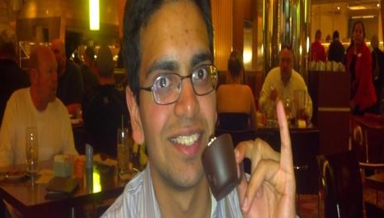}}\hskip.2em
    \subfloat[]{\includegraphics[width=0.23\textwidth]{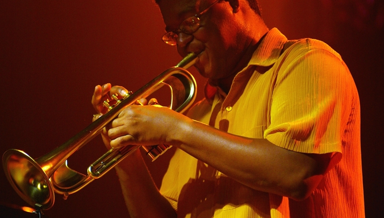}}
    \vspace{-.2cm}
  \caption{Images with approximately the same linearly re-scaled MOS exhibit dramatically different perceptual quality. If the human annotations are in the form of difference MOSs (DMOSs), we first negate the values followed by linear re-scaling. They are sampled from: (a) LIVE~\cite{sheikh2006statistical}, (b) CSIQ~\cite{larson:011006}, (c) TID2013~\cite{Ponomarenko201557}, (d) BID~\cite{ciancio2011no}, (e) LIVE Challenge~\cite{ghadiyaram2016massive}, and (f) KonIQ-10K~\cite{koniq10k}.}
\label{fig:quality_comparison}
\end{figure}

Very limited effort has been put to develop {\em unified} BIQA models for both synthetic and realistic distortions. Mittal~\textit{et al.}~\cite{mittal2013making} based their NIQE method on a prior probability model of natural undistorted images, aiming for strong generalizability to unseen distortions. However, NIQE is only able to handle a small set of synthetic distortions. Zhang~\textit{et al.}~\cite{zhang2015feature} extended NIQE~\cite{mittal2013making} by extracting more powerful natural scene statistics for local quality prediction. Another seemingly plausible solution is to directly combine multiple IQA databases for training. However, existing databases have different perceptual scales due to differences in subjective testing methodologies (see Table~\ref{tab:database}). A separate subjective experiment on images sampled from each database is then required  for perceptual scale realignment~\cite{sheikh2006statistical,larson:011006}. To emphasize this point, we linearly re-scale the mean opinion scores (MOSs) of each of the six databases to $[0,100]$, and show sample images that have approximately the same re-scaled MOS in Fig.~\ref{fig:quality_comparison}. It is clear that  they appear to have dramatically different perceptual quality (as expected). Using the noisy re-scaled MOSs for training results in suboptimal performance (see Table~\ref{tab:overall20}).

\textbf{\begin{table}[t]
  \small
  \centering
  \caption{Comparison of subject-rated IQA databases. MOS stands for mean opinion score. DMOS is inversely proportional to MOS. }\label{tab:database}
  \begin{tabular}{l|ccc}
      \toprule
        {Database} & Scenario & Annotation  & Range\\
     \hline
        LIVE~\cite{sheikh2006statistical} & Synthetic & DMOS & [0, 100]\\
        CSIQ~\cite{larson:011006} & Synthetic & DMOS & [0, 1]\\
        TID2013~\cite{Ponomarenko201557} & Synthetic & MOS & [0, 9]\\
    \hline
        BID~\cite{ciancio2011no} & Realistic & MOS & [0, 5]\\
        LIVE Challenge~\cite{ghadiyaram2016massive} &  Realistic & MOS & [0, 100]\\
        KonIQ-10K~\cite{koniq10k} & Realistic & MOS & [1, 5]\\
     \bottomrule
   \end{tabular}
\end{table}}

In addition to training with \textit{absolute} MOSs, recent methods also exploit \textit{relative} ranking information to learn BIQA models from three major sources: distortion specifications~\cite{liu2017rankiqa, ma2018end}, full-reference IQA models~\cite{ye2014beyond,ma2017dipiq,ma2019blind}, and human data~\cite{gao2015learning}. Liu~\textit{et al.}~\cite{liu2017rankiqa} and Zhang~\textit{et al.}~\cite{zhang2019deep} extracted ranking information from images of the same content and distortion type but different levels to pre-train deep neural networks (DNNs) for subsequent quality prediction. Their methods can only be applied to synthetic distortions, whose degradation processes are exactly specified. Ma~\textit{et al.}~\cite{ma2017dipiq,ma2019blind} supervised the learning of BIQA models with ranking information from full-reference IQA models. Their methods cannot be extended to realistic distortions either because the reference images are not available or may not even exist for full-reference models to compute quality values. The closest work to ours is due to Gao~\textit{et al.}~\cite{gao2015learning}, who inferred binary ranking information from MOSs. However, they neither performed joint optimization of feature extraction and quality prediction in an end-to-end fashion nor explored the idea of combining multiple IQA databases via pairwise rankings. Therefore,  their model only delivers reasonable performance on a small set of synthetic distortions. We summarize and compare previous ranking-based BIQA methods in Table~\ref{tab:rank_comparison}.

\textbf{\begin{table*}[t]
  \small
  \centering
  \caption{Summary of ranking-based BIQA models. DS: distortion specification characterized by distortion parameters. FR: full-reference IQA model predictions. std: standard deviation.}\label{tab:rank_comparison}
  \begin{tabular}{l|ccccccc}
      \toprule
        {Model} & RankIQA~\cite{liu2017rankiqa} & DB-CNN~\cite{zhang2019deep} & dipIQ~\cite{ma2017dipiq} & Ma~\textit{et al.}~\cite{ma2019blind} & Gao~\textit{et al.}~\cite{gao2015learning} & Ours\\
     \hline
        Source & DS & DS & FR & FR & (D)MOS & (D)MOS+std\\
        Scenario & Synthetic & Synthetic & Synthetic & Synthetic & Synthetic & Synthetic+Realistic\\
        Annotation & Binary & Categorical & Binary & Binary & Binary & Continuous\\
        Loss & Hinge variant & Cross entropy & Cross entropy & Cross entropy variant & Hinge & Fidelity\\
     \bottomrule
   \end{tabular}
\end{table*}}


In this paper, we aim to develop a \textit{unified} BIQA model for both synthetic and realistic distortions (with a single set of model parameters). To achieve this, we describe a novel training strategy, which involves two steps: IQA database combination and pairwise learning-to-rank model estimation. First, we build a training set by combining image pairs sampled within each individual IQA database (\ie, in the intra-database setting), which bypasses additional subjective experiments for perceptual scale realignment. Under the Thurstone's model~\cite{thurstone1927law}, a \textit{continuous} quality annotation for each pair can be computed  using the corresponding MOSs and the standard deviations (stds), which indicates the \textit{probability} of an image having higher perceived quality than the other. Comparing to a \textit{binary} quality label that identifies which one of an image pair has better quality~\cite{gao2015learning,ma2019blind},  our continuous quality annotation provides a more informative and reliable measurement of relative perceptual quality. We learn a DNN for BIQA using a pairwise learning-to-rank technique with the fidelity loss~\cite{tsai2007frank}. 
Although our training set does not include image pairs across different databases (\ie, in the inter-database setting~\cite{perez2020pairwise}), experimental results on six databases covering both synthetic and realistic distortions suggest that the trained BIQA method on massive image pairs generated in the intra-database setting outperforms existing BIQA models by a large margin.

\section{Method}
\label{sec:ubiqe}
In this section, we present in detail the proposed training strategy consisting of IQA database combination and pairwise learning-to-rank model estimation, followed by the network specification.

\subsection{Training Set Construction}\label{training_set}
Given $m$ subject-rated IQA databases, we randomly sample from the $j$-th database $n_j$ image pairs $\lbrace( x^{j}_i,y^{j}_i)\rbrace_{i=1}^{n_j}$. For each pair $( x^{j}_i, y^{j}_i)$, we infer its relative ranking information from the corresponding MOSs and stds. Specifically, we make use of the Thurstone's model~\cite{thurstone1927law} and assume that the true perceptual quality $q(x)$ follows a Gaussian distribution with mean $\mu(x)$ ({\em i.e.}, the MOS) and std $\sigma(x)$ collected via subjective testing. The quality difference is also Gaussian with mean $\mu(x) - \mu(y)$ and std $\sqrt{\sigma^{2}(x) + \sigma^{2}(y)}$, assuming independence between $x$ and $y$. The probability (denoted by $p(x,y)\in [0,1]$) that $x$ has higher perceptual quality than $y$ can be computed by
\begin{align}\label{eq:difference}
p(x,y)= \Pr({q(x) \ge q(y)}) = \Phi\left(\frac{\mu(x) - \mu(y)}{\sqrt{\sigma^{2}(x) + \sigma^{2}(y)}}\right),
\end{align}
where $\Phi(\cdot)$ is the Normal cumulative distribution function. By combining image pairs from $m$ IQA databases, we are able to construct a training set $\mathcal{D}= \lbrace\lbrace(x^j_i,  y^j_i),p^j_i\rbrace_{i=1}^{n_j}\rbrace_{j=1}^{m}$. A significant advantage of our database combination approach is that it does not require any subjective realignment experiment, allowing future IQA databases to be added in constructing $\mathcal{D}$ with essentially no cost.

\subsection{Model Estimation}\label{L2R}
Given the training set $\mathcal{D}$, our goal is to learn two differentiable functions $f_{w}(\cdot)$ and $\sigma_{w}(\cdot)$, parameterized by a vector $w$, which take an image $x$ as input, and compute the  quality prediction value  and its uncertainty. Similar in Section~\ref{training_set}, we assume the true perceptual quality $q(x)$ obeys a Gaussian distribution with mean and std now estimated by $f_w(x)$ and $\sigma_w(x)$, respectively. The probability (denoted by $p_w(x,y)\in[0,1]$) that $x$ has better perceived quality than $y$ in an image pair can be estimated by 
\begin{align}\label{eq:difference2}
p_w(x,y)=\Pr({q(x) \ge q(y); w}) = \Phi\left(\frac{f_{w}(x) - f_{w}(y)}{\sqrt{\sigma_{w}^{2}(x) + \sigma_{w}^{2}(y)}}\right).
\end{align}
While general similarity measures between probability distributions such as cross entropy and Kullback-Leibler divergence can be used as the criteria for model estimation, they suffer from several problems~\cite{tsai2007frank}. First, the minimal value of the cross entropy loss for an image pair $(x,y)$ with the ground truth $0<p(x,y)<1$ ({\em i.e.,} other than zero and one) is not exactly zero, which may hinder the learning procedure (see Table~\ref{tab:overall}). Second, the cross entropy loss is unbounded from above, which may give excessive penalties to (and therefore bias towards) some hard training examples. To address the above problems,  we adopt the fidelity loss~\cite{tsai2007frank}, originated from quantum physics to measure the difference between two states of a quantum~\cite{birrell1984quantum} as our objective function
\begin{align}\label{eq:fidelity}
\ell(x, y, p;w)
=& 1 - \sqrt{p(x,y)p_w(x,y)}  \nonumber \\
&-\sqrt{(1-p(x,y))(1-p_w(x,y))}.
\end{align}
In practice, we sample a mini-batch $\mathcal{B}$ from $\mathcal{D}$ in each iteration and use a variant of the stochastic gradient descent method to adjust the parameter vector $w$ by minimizing the following empirical loss
\begin{align}\label{eq:loss1}
\ell(\mathcal{B}; w) =\frac{1}{\vert\mathcal{B}\vert} \sum_{\{( x, y), p\}\in \mathcal{B}}\ell( x,  y,p; w),
\end{align}
where $\vert\mathcal{B}\vert$ represents the cardinality of $\mathcal{B}$. 

\subsection{Network Specification}\label{ranknet}
\begin{figure}
  \centering
  \includegraphics[width=.48\textwidth]{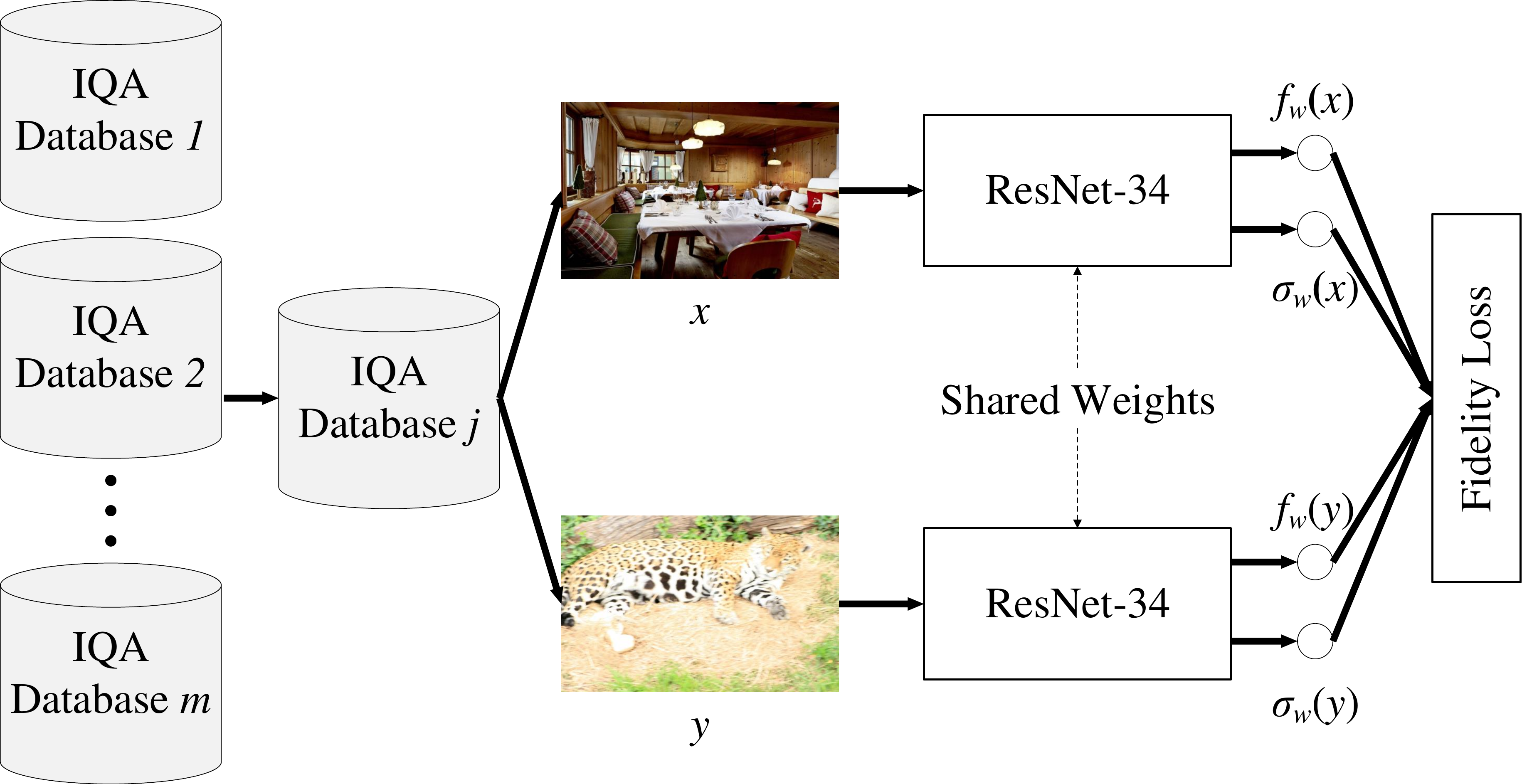}
  \caption{The Siamese  framework for learning the quality prediction function $f_{w}(\cdot)$ and its associated uncertainty $\sigma_{w}(\cdot)$, driven by the fidelity loss. The training image pairs are randomly sampled within individual IQA databases. }\label{fig:fig1}
\end{figure}

Due to the success of DNNs in various computer vision and image processing applications, we adopt a ResNet~\cite{he2016deep} as the backbone to construct our quality prediction function $f_{ w}(x)$ and estimate the uncertainty $\sigma_w(x)$. The Siamese  learning framework consisting of two streams is shown in Fig.~\ref{fig:fig1}. Each stream is composed of a stage of convolution, batch normalization~\cite{ioffe2015batch}, ReLU nonlinearity, and max-pooling, followed by four groups of layers based on the bottleneck architecture~\cite{he2016deep}. To better summarize spatial statistics and generate a fixed-length representation regardless of input image size, we replace the first-order average pooling in the original ResNet with a second-order bilinear pooling~\cite{zhang2019deep}, which has been proven to be effective in object recognition~\cite{lin2015bilinear} and BIQA~\cite{zhang2019deep}. Denoting the spatially flattened feature representation after the last convolution by $z \in \mathbb{R}^{s \times c}$, where $s$ and $c$ denote the spatial and channel dimensions, respectively, we define the bilinear pooling  as
\begin{align}\label{eq:bp}
\bar{z} = z^T z.
\end{align}
We further flatten $\bar{z}\in \mathbb{R}^{c\times c}$ and append a fully connected layer to compute two scalars that represent the perceptual quality and its uncertainty. The weights of the two streams are shared  during the entire optimization process.

\begin{table*}[t]
  \centering
  \caption{Median SRCC results across ten sessions on the \underline{test sets}  of the six IQA databases covering both synthetic and realistic distortions. The training databases for models (relying on human annotations) are highlighted in the bracket.}\label{tab:overall20}

  \begin{tabular}{l|cccccc}
      \toprule
     & {LIVE~\cite{sheikh2006statistical}} & {CSIQ~\cite{larson:011006}} & {TID2013~\cite{Ponomarenko201557}} & {BID~\cite{ciancio2011no}} & {LIVE Challenge~\cite{ghadiyaram2016massive}} & {KonIQ-10K~\cite{koniq10k}}\\
    \hline
       MS-SSIM~\cite{wang2003multiscale} & 0.951 & 0.910 & 0.790 & -- & -- & -- \\
       NLPD~\cite{Laparra:17} & 0.942 & 0.937 & 0.798 & -- & -- & -- \\
     \midrule
        NIQE~\cite{mittal2013making} & 0.906 & 0.632 & 0.343 & 0.468 & 0.464 & 0.521 \\
       ILNIQE~\cite{zhang2015feature} & 0.907 & 0.832 & 0.658 & 0.516 & 0.469 & 0.507 \\
       dipIQ~\cite{ma2017dipiq} & 0.940 & 0.511 & 0.453 & 0.009 & 0.187 & 0.228\\
    \hline
      MEON (LIVE)~\cite{ma2018end}  & -- & 0.726 & 0.378 & 0.100 & 0.378 & 0.145 \\
      deepIQA (TID2013)~\cite{bosse2018deep} & 0.833 & 0.687 & -- & 0.120 & 0.133 & 0.169 \\
      PQR (BID)~\cite{zeng2017probabilistic} & 0.634 & 0.559 & 0.374 & -- & 0.680 & 0.636 \\
      PQR (KonIQ-10K)~\cite{zeng2017probabilistic} & 0.729 & 0.709 & 0.530 & 0.755 & 0.770 & -- \\
       DB-CNN (TID2013)~\cite{zhang2019deep} & 0.883 & 0.817 & -- & 0.409 & 0.414 & 0.518 \\
       DB-CNN (LIVE Challenge)~\cite{zhang2019deep} & 0.709 & 0.691 & 0.403 & 0.762 & -- & 0.754 \\
        \hline
       Linear re-scaling (All databases) & 0.924 & 0.807 &  0.746  & 0.842  & 0.824 & 0.880 \\
       Binary labeling (All databases)& {\bf 0.957} & {\bf 0.867} & {\bf 0.806} & {\bf 0.851} & {\bf 0.853} & {\bf 0.892} \\
     \hline
       Ours (All databases)& {\bf 0.961} & {\bf 0.907} & {\bf 0.855} & {\bf 0.863} & {\bf 0.851} & {\bf 0.894} \\
     \bottomrule
   \end{tabular}
\end{table*}

\begin{table}[t]
  \centering
  \caption{SRCC results on the four IQA databases under the cross-database setup. DB-CNN$_s$ and DB-CNN$_r$ stand for the DB-CNN method trained on TID2013~\cite{Ponomarenko201557} and LIVE Challenge~\cite{ghadiyaram2016massive}, respectively. Our method is trained on TID2013 and LIVE Challenge simultaneously. }\label{tab:overall}

  \begin{tabular}{l|cccc}
      \toprule
    Database & {LIVE} & {CSIQ} & {BID} & {KonIQ-10K}\\
    \hline
     NIQE & 0.906 & 0.627 & 0.459 & 0.530 \\
     ILNIQE & 0.898 & \textbf{0.815} & 0.494 & 0.506 \\
     dipIQ & \textbf{0.938} & 0.527 & 0.019 & 0.238 \\
     DB-CNN$_s$ & 0.903 & 0.769 & 0.434 & 0.516 \\
     DB-CNN$_r$ & 0.746 & 0.697 & \textbf{0.832} & \textbf{0.776} \\
    \hline
     Ours & \textbf{0.921} & \textbf{0.821} & \textbf{0.840} & \textbf{0.794} \\

     \bottomrule
   \end{tabular}
\end{table}

\section{Experiments}
\label{sec:exp}
In this section, we first describe training and testing procedures. We then compare the performance of our method to a set of state-of-the-art BIQA models.

\subsection{Model Training}\label{protocol}
We generate the training set and conduct comparison experiments on six IQA databases, including LIVE~\cite{sheikh2006statistical}, CSIQ~\cite{larson:011006}, TID2013~\cite{Ponomarenko201557}, LIVE Challenge~\cite{ghadiyaram2016massive}, BID~\cite{ciancio2011no}, and KonIQ-10K~\cite{koniq10k}. The first three are synthetically distorted, while the last three are realistically distorted. More information of these databases can be found in Table~\ref{tab:database}. We randomly sample $80\%$ of the images in each database for training and leave the rest for evaluation. Regarding LIVE, CSIQ, and TID2013, we split training and test sets according to the reference images such that content independence between the two sets is guaranteed. From the six databases, we are able to generate more than $240,000$ image pairs.
During testing, we use Spearman rank-order correlation coefficient (SRCC) to quantify the performance on individual databases.


We adopt ResNet-34~\cite{he2016deep} as the backbone of $f_{w}(x)$ and $\sigma_{w}(x)$, and train it using the Adam optimizer~\cite{Kingma2014adam} for twelve epochs. The initial learning rate is set to  $10^{-4}$ with a decay factor of $10$ for every three epochs. A warm-up training strategy is adopted: only the last fully connected layer with random initialization is learned in the first three epochs with a mini-batch of $128$; for the remaining epochs, we fine-tune the entire network with a mini-batch of $32$. 
In all experiments, we test on images of original size. To reduce the bias caused by the randomness in training and test set splitting, we repeat this process for \underline{ten} times and report the median SRCC results.

\subsection{Model Performance}\label{results}
\noindent\textbf{Main Results}. We compare our method with three knowledge-driven BIQA models that do not require MOSs for training - NIQE~\cite{mittal2013making}, ILNIQE~\cite{zhang2015feature} and dipIQ~\cite{ma2017dipiq}, and four data-driven DNN-based models -  MEON~\cite{ma2018end}, deepIQA~\cite{bosse2018deep}, PQR~\cite{zeng2017probabilistic} and DB-CNN~\cite{zhang2019deep}. The implementations are obtained from the respective authors. The results are listed in Table~\ref{tab:overall20}, where we have several interesting observations. First, our method significantly outperforms the three knowledge-driven models. Although NIQE~\cite{mittal2013making} and its feature-enriched version ILNIQE~\cite{zhang2015feature} are designed for arbitrary distortion types, they do not perform well on realistic distortions and challenging synthetic distortions in TID2013~\cite{Ponomarenko201557}. dipIQ~\cite{ma2017dipiq} is only able to handle distortion types that have been seen during training. As a result, the performance of dipIQ on all databases except LIVE is particularly weak,  highlighting the difficulties of distortion-aware BIQA methods to handle unseen distortions.

We then compare our model with four recent DNN-based methods. Since previous data-driven models can only be trained on one IQA database, we highlight in the table the databases used to train the respective models. Despite pre-trained on a large number of synthetically distorted images, MEON fine-tuned on LIVE~\cite{sheikh2006statistical} does not generalize to other databases with different distortion types and scenarios. Although trained with more synthetic distortion types in TID2013~\cite{Ponomarenko201557}, deepIQA~\cite{bosse2018deep} performs slightly worse on CSIQ~\cite{larson:011006} and LIVE Challenge~\cite{ghadiyaram2016massive} than MEON due to label noise in patch-based training. By bilinearly pooling two feature representations that are sensitive to synthetic and realistic distortions, respectively, DB-CNN~\cite{zhang2019deep} trained on TID2013 achieves reasonable performance on databases built in the wild. Based on a probabilistic formulation, PQR~\cite{zeng2017probabilistic} aims for realistic distortions, and trains two models on BID~\cite{ciancio2011no} and KonIQ-10K~\cite{koniq10k} separately. Benefiting from a larger number of training images with more diverse content and distortion variations, PQR trained on KonIQ-10K generalizes much better to the rest databases than the one trained on BID. Our method performs significantly better than all competing models on all six databases, and is close to two full-reference models (MS-SSIM~\cite{wang2003multiscale} and NLPD~\cite{Laparra:17}). We believe this performance improvement arises because the proposed learning technique allows us to train the proposed method on multiple databases simultaneously, adapting feature representations to multiple distortion scenarios. In addition, the pre-trained weights on object recognition are helpful to prevent our method over-fitting to any single pattern within individual databases.

Finally, we test the proposed training strategy in a more challenging cross-database setting. Specifically, we construct another training set $\mathcal{D}'$ using image pairs sampled from the full TID2013~\cite{Ponomarenko201557} and LIVE Challenge~\cite{ghadiyaram2016massive} databases, and re-train the proposed model. We compare its performance against the state-of-the-art on LIVE~\cite{sheikh2006statistical}, CSIQ~\cite{larson:011006}, BID~\cite{ciancio2011no},  and KonIQ-10K~\cite{koniq10k}. As can be seen from Table~\ref{tab:overall}, our model delivers significantly better performance than the three knowledge-driven models and the best-performing DNN-based model DB-CNN (trained on TID2013 and LIVE Challenge separately). This provides strong evidence that our optimized method by the proposed training strategy generalizes to both synthetic and realistic distortion scenarios. The training image pairs from the two databases effectively provide mutual regularization, guiding the network
to a better local optimum.

\noindent\textbf{Ablation Study}. We first train a baseline model on the six IQA databases using the linearly re-scaled MOSs. As listed in Table~\ref{tab:overall20}, the performance of the baseline model drops significantly due to the noise introduced by linear re-scaling (see Fig.~\ref{fig:quality_comparison}). To verify the effectiveness of the fidelity loss, we replace it with the cross entropy loss and re-train our model. For an image pair $(x, y)$, the ground truth binary label $r=1$, if $u(x)\ge u(y)$, indicating that $x$ is of higher quality, and  otherwise $r=0$. Comparing to the method optimized for the cross entropy loss in Table~\ref{tab:overall20}, the proposed model optimized for the fidelity loss achieves comparable performance on realistically distorted databases, but is significantly better on synthetically distorted databases. 

\section{Conclusion and Discussion}
\label{sec:conclusion}
We have introduced a BIQA model and a method of training it on multiple IQA databases. Our BIQA model is the first of its kind to deliver superior performance on both synthetically and realistically distorted  databases with a single set of model parameters. The proposed learning strategy is model agnostic, meaning that it can be combined with other data-driven BIQA models, especially advanced ones for improved performance. In addition, it is straightforward to incorporate more image pairs into training, when new IQA databases are available. We hope that the proposed learning strategy will become a standard solution for existing and next-generation BIQA models to meet the cross-distortion-scenario challenge.

In the future, we plan to explicitly enforce additional constraints when learning the uncertainty by exploiting the available ground truth stds, towards a more sensible uncertainty-aware BIQA model. It will also be important to compare the generalizability of our method more thoroughly to recent DNN-based BIQA models using the group maximum differentiation competition methodology~\cite{ma2019group} on a large-scale image database of both synthetic and realistic distortions.


\bibliographystyle{IEEEbib}
\bibliography{weixia}

\end{document}